# AAE 568 Course Project Report
# Optimal Control for Constrained Coverage Path Planning

Ankit Manerikar, Debasmit Das, Pranay Banerjee

August 9, 2017

## ABSTRACT

The problem of constrained coverage path planning involves a robot trying to cover maximum area of an environment under some constraints that appear as obstacles in the map. Out of the several coverage path planning methods, we consider augmenting the linear sweep-based coverage method to achieve minimum energy/ time optimality along with maximum area coverage. In addition, we also study the effects of variation of different parameters on the performance of the modified method.

## INTRODUCTION

Most Coverage Path Planning algorithms rely on breaking down the free space(i.e the obstacle-free space) into simple, non-overlapping sub-regions called cells. A survey of the coverage path planning problem is given in [3]. Two of the most popular offline (environment assumed to be known *a priori*) cellular decomposition approaches are the **trapezoidal** decomposition and the **boustrophedon** decomposition. The trapezoidal decomposition technique [1] divides the free space into trapezoidal cells, and each cell, having two parallel sides, can be covered by simple back and forth motions parallel to either side called *slices* with the sweep direction being between the non-parallel sides. Therefore, coverage is ensured by visiting each cell in the adjacency graph. The shortcoming of this method is that it requires far too much redundant back and forth motions to guarantee complete coverage. The boustrophedon decomposition [2] compensates for the redundant movements by merging the cells that do not contribute to change in connectivity of the nodes in the adjacency graph. This merging technique reduces the number of cells in the decomposition, thereby, reducing the overall number of back and forth motions. However, all these methods rely on maximizing the area covered without considering the time and energy spent. We introduce these additional requirements in our project.

In our project, we have modeled the mobile robot as a **point** that can move on a **2D plane**. The robot has a coverage represented as a circle centered at the location of the robot and a radius that represents the extent of coverage. The goal is to cover the entire area in **minimum time** and by expending **minimum energy**. We can make the problem more challenging by including obstacles. We will primarily keep the shape of obstacles as circular. We will introduce various constraints other than that of the system dynamics. Firstly, the point robot should not move beyond the area constraints. Secondly, the point robot should not move into the obstacle region. We allow the mobile robot to move on the boundary of the area or the boundary of the obstacle. We will represent the performance criterion, the system dynamics and the constraints mathematically in the subsequent sections.

## MAIN BODY

The problem of coverage path planning can be visualized as shown in Figure 1. The point robot has to cover the whole area of the given 2D surface. The point robot's coverage is shown in grey, while the obstacle is shown in black. Based on the path history of the point-robot, the covered area is shown as a bold dotted line. We would like the robot to **cover maximum area** possible on the 2D surface. However, there maybe intersecting trajectories such that the area covered overlaps. So, we would also like to **minimize the area overlapped**. Unfortunately, the area covered and the area overlapped are difficult to be represented mathematically, so we use heuristic techniques from the literature. The minimum time-energy problem is solved using optimal control. We will first discuss heuristic based techniques for area coverage. The problem described above can have many practical applications like floor cleaning, agricultural inspection etc. which are discussed in detail in [3].

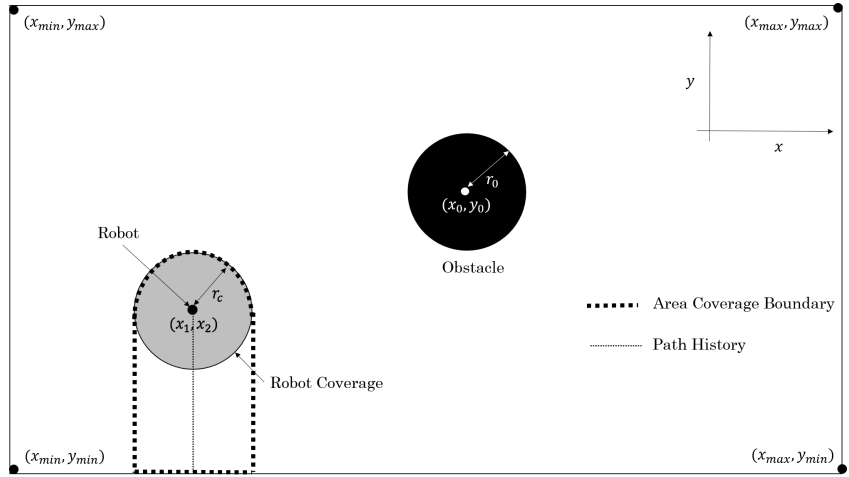

Figure 1: Complete visualization of our problem

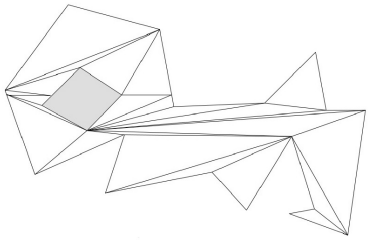

Figure 2: Complete triangularization (Courtesy [5])

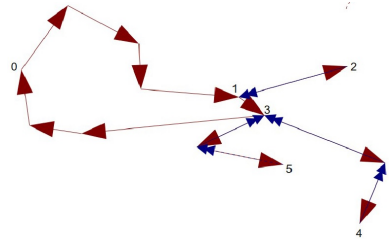

Figure 3: Hamiltonian Virtual Circuit (Courtesy [5])

## Heuristic Methods

There are several approaches to coverage path planning that ensure maximization of the coverage area, however, not all of them guarantee minimum area overlap. The approach of convex decomposition [5], for example, transforms the real area into an approximate representation of connected linear segments. This approach defines the obstacles as holes but with a sequence of linear segments connected in inverse order compared to the outer layer. This is followed by triangularization, which divides each element into triangular polygons. The result of complete triangularization is shown in Figure 2 Creating a connected graph between the centers of each polygon leads to the Hamiltonian circuit [6] shown in Figure 3. It consists of virtual connections(shown with double head arrows), which are connections between any directly unconnected pair, through intermediate vertexes via path overlapping. The direction of navigation in the circuit of the graph is determined by the arrows. With the circuit now defined, and the entry and exit edges already selected, a coverage path algorithm can be generated. However, due to the virtual connections in the circuit, the algorithm would not satisfy the minimum overlap constraint in our problem.

An alternative approach to coverage path algorithms is the following: decomposing the coverage region into sub-regions, generating a sequence of sub-regions to visit and creating a coverage path from this sequence that covers each sub-region in turn. However, unlike the previous case, these algorithms all use a single line sweep in order to decompose the coverage region into sub-regions, and these sub-regions are individually covered using a back and forth motion in rows perpendicular to the sweep direction. All sub-regions use the same sweep direction.

One such approach that explicitly performs a line sweep decomposition and creates a sequence of sub-regions (cells) using an heuristic Traveling Salesman algorithm is the so-called Boustrophedon decomposition proposed by Choset and Pignon [3]. This decomposition method is designed in such a way as to minimize the number of excess lengthwise motions. The Boustrophedon decomposition method uses the idea of exact cellular decomposition, which is a motion planning technique in which the free configuration space (set of all robot configurations where the robot does not overlap an obstacle) is decomposed into cells such that the union of the cells is the original free space. This decomposition approach assumes that a vertical line, termed a slice, sweeps left to right through a bounded environment which is populated with polygonal obstacles. Cells are formed via a sequence of open and close operations which occur when the slice encounters an event, an instance in which a slice intersects a vertex of a polygon. The IN event corresponds to the slice intersecting the first vertex of the polygonal obstacle, which increases the connectivity from one to two, thereby leading to the formation of two new cells, as shown in Figure 4. The OUT event corresponds to the slice intersecting the last vertex of the polygonal obstacle, changing the connectivity from two to one, thereby merging two cells into one cell, as shown in Figure 5.

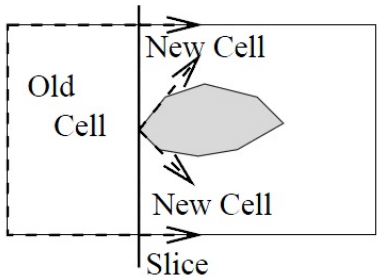 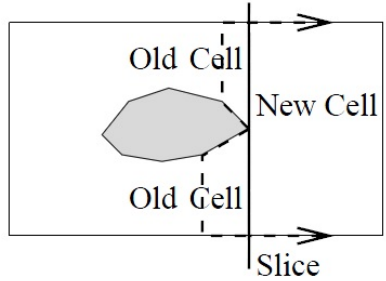 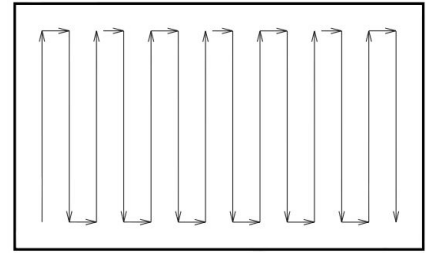

Figure 4: IN event    Figure 5: OUT event    Figure 6: Boustrophedon Path

Once the decomposition and adjacency graph are determined, a mobile robot employs a simple graph search algorithm to determine a walk through the adjacency graph that visits all nodes, i.e., visits all cells. Since simple back-and-forth motions covers each cell, complete coverage is achieved by visiting each cell, as shown in Figure 6. The Boustrophedon decomposition ensures complete coverage with no area overlap, however, it does not ensure minimization of time or energy. Huang [4] adapts the Boustrophedon approach to achieve optimal coverage in terms of time and energy. Huang shows that the optimal line sweep decomposition must use a sweep line that is parallel to an edge of the boundary or an obstacle or the convex hull for a polygonal environment, minimizing the number of turns taken for complete coverage, thereby optimizing time and energy.

Thus, ensuring that the mobile robot follows the Boustrophedon decomposition with Huang's optimality criterion would result in a complete coverage with minimum area overlap, minimum time and minimum energy expenditure. So the problem of coverage path planning is reduced to finding a minimum time/energy trajectory for each slice.

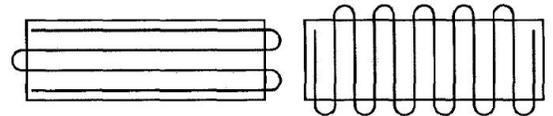

## Optimal Control

Figure 7: Number of turns influences coverage cost(Courtesy [4])

We formulate the **optimal control problem for each slice**. The initial and final states are fixed and the final time is free. We will try to dualize the problem into a (TPBVP) two-point boundary value problem using the state constraints and the obstacle. For the analysis we will use an example of 1 obstacle -

$$J = \int_0^{t_f} \{\underbrace{(1-w)(u_1^2 + u_2^2)}_{\text{Energy}} + \underbrace{w}_{\text{Time}}\} dt \quad ; \quad 0 \leq w \leq 1 \tag{1}$$

with Running Cost as the bracketed integrand.

$$\overbrace{\begin{bmatrix} \dot{x}_1 \\ \dot{x}_2 \\ \dot{x}_3 \\ \dot{x}_4 \end{bmatrix}}^{\dot{\mathbf{x}}} = \overbrace{\begin{bmatrix} 0 & 0 & 1 & 0 \\ 0 & 0 & 0 & 1 \\ 0 & 0 & 0 & 0 \\ 0 & 0 & 0 & 0 \end{bmatrix}}^{A} \overbrace{\begin{bmatrix} x_1 \\ x_2 \\ x_3 \\ x_4 \end{bmatrix}}^{\mathbf{x}} + \overbrace{\begin{bmatrix} 0 & 0 \\ 0 & 0 \\ 1 & 0 \\ 0 & 1 \end{bmatrix}}^{B} \overbrace{\begin{bmatrix} u_1 \\ u_2 \end{bmatrix}}^{\mathbf{u}} \quad ; \quad \mathbf{x}(0) = \begin{bmatrix} 0 \\ 0 \\ 0 \\ 0 \end{bmatrix} \quad ; \quad \mathbf{x}(t_f) = \begin{bmatrix} 10 \\ 0 \\ 0 \\ 0 \end{bmatrix} \quad (2)$$

$$0 \leq x_1 \leq 10; 0 \leq x_2 \leq 2r(r = 0.1); x_1 \geq 0; 10 - x_1 \geq 0; x_2 \geq 0; 10 - x_2 \geq 0 \quad (3)$$

$$(x_1 - x_0)^2 + (x_2 - y_0)^2 \geq r_o^2; \quad (4)$$

We formulate the Hamiltonian for the necessary conditions of optimality with state inequalities as follows (using the concept of augmented state) :

$$\begin{aligned} H = & w + (1-w)(u_1^2 + u_2^2) + \lambda_1 x_3 + \lambda_2 x_4 + \lambda_3 u_1 + \lambda_4 u_2 + \lambda_5[(x_1)^2 \mathbb{1}(-x_1) + (10-x_1)^2 \mathbb{1}(x_1 - 10) + \\ & (x_2)^2 \mathbb{1}(-x_2) + (10-x_2)^2 \mathbb{1}(x_2 - 10) + \{(x_1 - x_0)^2 + (x_2 - y_0)^2 - r_o^2\}^2] \\ & \mathbb{1}\{r_o^2 - (x_1 - x_0)^2 + (x_2 - y_0)^2\}] \end{aligned} \quad (5)$$

Now, the necessary conditions are given as follows-

$$\dot{x}_1^* = \frac{\partial H}{\partial \lambda_1} \quad ; \quad \dot{x}_2^* = \frac{\partial H}{\partial \lambda_2} \quad ; \quad \dot{x}_3^* = \frac{\partial H}{\partial \lambda_3} \quad ; \quad \dot{x}_4^* = \frac{\partial H}{\partial \lambda_4} \quad ; \quad (6)$$

These are just the state equations Also the costate equations are given by-

$$\dot{\lambda}_1^* = -\frac{\partial H}{\partial x_1} \quad ; \quad \dot{\lambda}_2^* = -\frac{\partial H}{\partial x_2} \quad ; \quad \dot{\lambda}_3^* = -\frac{\partial H}{\partial x_3} \quad ; \quad \dot{\lambda}_4^* = -\frac{\partial H}{\partial x_4} \quad ; \quad \dot{\lambda}_5^* = 0 \quad ; \quad (7)$$

Also, using the principle to find for $u_1^*$ and $u_2^*$ -

$$H(\mathbf{x}^*, \mathbf{u}^*, \boldsymbol{\lambda}^*, t) \leq H(\mathbf{x}^*, \mathbf{u}, \boldsymbol{\lambda}^*, t) \quad (8)$$

Now, since the final time $t_f$ is not free and the Hamiltonian does not explicitly depend on time-

$$H(\mathbf{x}^*, \mathbf{u}^*, \boldsymbol{\lambda}^*) = 0 \quad (9)$$

Equations (6), (7), (8), (9) need to hold for all $t \in [0, t_f]$. From the above equations, we can formulate a Two-Point Boundary Value Problem (TPBVP) that we feed into an ODE Solver, say `bvp4c()` in MATLAB to get the optimal state and control trajectories. However, as we increase the number of obstacles, the necessary conditions become complex and we have to manually solve for these conditions everytime and feed into the ODE Solver. Moreover, the solution to the TPBVP is very sensitive to the initial guess. Rather we look at a relatively modern technique of Pseudospectral Optimal Control and discuss that in context with our problem in hand. Pseudospectral theory has the following advantages - 1) **Fast Convergence** irrespective of number of obstacles. 2) Alleviate **Curse of Sensitivity** (Not Sensitive to initial guess). 3) Implementation is easy.

### Collocation Methods

There are number of collocation methods that can be used to discretize the time-interval into $N$ steps. It is to be noted that the $N$ time-intervals may not be uniform. Also, we do not expect the state constraints to follow the dynamics within the interpolated time. The collocation method used depends upon our problem at hand. Since the initial and final states are specified we follow the Legendre-Gauss-Lobato (LGL) protocol for interpolation at the $N$ points. A detailed description on collocation is given in [7]. As for the pseudospectral theory part, we need to shape the weighted interpolating functions represented by the states and the co-vectors. They are as follows -

$$x_k^N(t) = \sum_{j=0}^{N} \frac{W(t)}{W(t_j)} x_{kj} \phi_j(t) \quad ; \quad \lambda_k^N(t) = \sum_{j=0}^{N} \frac{W^*(t)}{W^*(t_j)} \lambda_{kj} \phi_j(t) \quad ; \quad u_k^N(t) = \sum_{j=0}^{N} u_{kj} \psi_j(t) \quad (10)$$

for all states $\{x_1, x_2, ...\}$, for all covectors $\{\lambda_1, \lambda_2, ...\}$ and for all controls $\{u_1, u_2...\}$. Here, $\phi_j(t)$ is the Lagrange Interpolating Polynomial [8] and $\psi_j(t)$ is a special interpolating function that makes the state and control trajectories dynamically feasible. $x_{kj}$ is the value of state $x_k$ at discrete time instant $t_j$. Similarly, $\lambda_{kj}$ is the value of the co-sate $\lambda_k$ at discrete time instant $t_j$. Since our problem requires LGL nodes, we would choose $W(t) = 1$ and $W(t_j) = 1$ respectively. This choice of the primal-dual weight functions have been mentioned in [9]. As referenced in Figure 8, we follow the direct approach where we first discretize the primal-problem using the interpolating functions. Therefore, the formulation of the problem $B^N$ in our context is given as follows -

**Cost Functional (Minimize) :**

$$J^N = \sum_{j=0}^{N} \{(1-w) + w(u_{1j}^2 + u_{2j}^2)\} \quad (11)$$

**State Equation :**

$$\sum_{j=0}^{N} D_{ij} x_{1j} = x_{3i} \quad ; \quad \sum_{j=0}^{N} D_{ij} x_{2j} = x_{4i} \quad ; \quad (12)$$

$$\sum_{j=0}^{N} D_{ij} x_{3j} = u_{1i} \quad ; \quad \sum_{j=0}^{N} D_{ij} x_{4j} = u_{2i} \quad ; \quad (13)$$

for all $i = 0, 1, ..N$ and $D_{ij}$ are differentiation matrix defined in [7].

**Boundary Conditions :**

$$x_{10} = 0 \ ; \ x_{20} = 0 \ ; \ x_{30} = 0 \ ; \ x_{40} = 0 \ ; \ x_{1N} = 10 \ ; \ x_{2N} = 0 \ ; \ x_{3N} = 0 \ ; \ x_{4N} = 0 \quad (14)$$

**Constraints :**

$$0 \leq x_{1i} \leq 10; 0 \leq x_{2i} \leq 2r(r=0.1) \quad (15)$$

$$(x_{1i} - x_0)^2 + (x_{2i} - y_0)^2 \geq r_o^2; \quad (16)$$

for all $i = 0, 1, 2...N$.

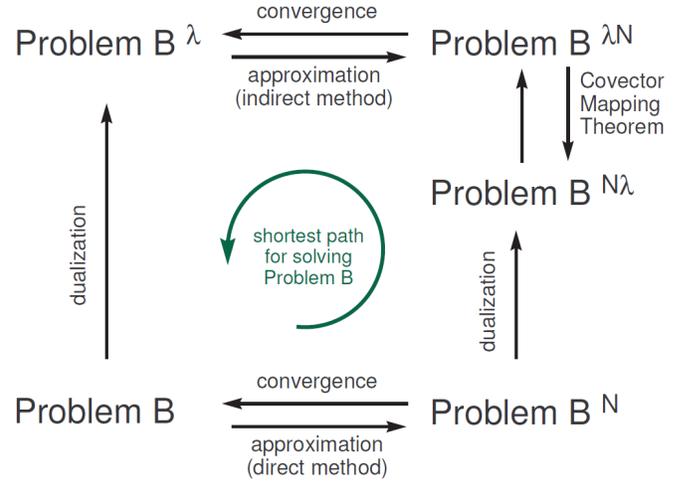

Figure 8: Optimal control cycle

Now the problem can be dualized to solve for the optimal state, costates and controls.

# IMPLEMENTATION

The reduction of the optimal coverage problem into a minimum time/minimum energy problem of the parallel sweep trajectories of a linear sweep-based coverage planning technique allows the implementation of an iterative algorithm that successively calculates the optimal path for the back-and-forth trajectory of the linear sweep for the given coverage area and in the presence of obstacles. The algorithm determines the optimum direction of sweep for the polygon based on the MSA (Minimum Sum of Altitudes) criterion [4] and updates the initial and final boundary values for the trajectory in each iteration depending upon the previous iteration that was calculated. An illustration of the implementation algorithm is given in Algorithm 1. From the above description, it can be seen that the determining the direction of sweep for the given area is a primary step not just in this algorithm but in any coverage planning technique. In this case, the optimum direction is calculated by using the MSA criterion which involves minimizing the following cost function:

$$\min \ S(\theta) = d_B(\theta) + \sum_i d_{H_i}(\theta) \quad (17)$$

**Algorithm 1** Algorithm for Optimal Sweep Trajectory Generation
───────────────────────────────────────────
1: **Start:**
2: **Input:**  $Coverage\_Area = \{\mathbf{e}_i \equiv (x_i, y_i), for\ i = 1, 2, ..., n\}$
3: $Obstacle\_Set = \{x_{obs}^{(i)}, y_{obs}^{(i)}, r_{obs}^{(i)}\},\ for\ i = 1, ..., N_{obs}\}$
4: Robot coverage, $r_{bot}$
5: **Output:** $Optimal\_Set, \{\mathbf{x}_i^*(t), for\ i = 1, ...\ N_{turn}\}$
6: **Variables:** State: $\mathbf{x} = (x_1, ..., x_4)^T$, Control: $\mathbf{u} = (u_1, u_2)^T$
7: **Get** $\theta^* = \min S(\theta) = d_B(\theta) + \sum_i d_{H_i}(\theta)$
8: **Select** Initial point, $\mathbf{x}_0$ along $\hat{\mathbf{n}}_{\theta^*}^{(0)}$
9: **for** k := 0 to $N_{turn}$ **do**
10: $Initial\_Guess : \{\mathbf{x}_0 = [tc_{\theta^*}\ ts_{\theta^*}\ v_{min}\ 0]^T\}$
11: Solve: min $J(\mathbf{x}(t), \lambda(t), \mathbf{u}(t), t)$
12: subject to $Coverage\_Area$
13: $Obstacle\_Set$
14: $\dot{\mathbf{x}}(t) = f(\mathbf{x}(t), \mathbf{u}(t)t)$
15: Set $\mathbf{x}_{k+1}(t_0) = \mathbf{x}_k^*(t_f) + 2r_{bot}.[c_{\theta^*},\ s_{\theta^*}]^T$
16: **end**
17: Final trajectory set = $\{\mathbf{x}_i^*(t), for\ i = 1, ...\ N_{turn}\}$
18: **Stop:**
───────────────────────────────────────────

where $d(\theta)$ is the diameter function, $H_i$ are the obstacles and $B$ is the boundary. This minimization ensures that the minimal number of turns are traversed by the coverage path as it sweeps the polygonal area. Once the direction is determined, the sweep lines are constructed normal to the the sweep direction by solving the minimum/energy optimal control problem. Note that after each iteration, the final position are used to determine the initial state of the next trajectory to be generated thereby generating the back-and-forth motion characteristic of the line-sweep decomposition methods. Because of such an iterative approach, the sweep trajectories remain parallel in the absence of obstacles hence inheriting the property of minimum area overlap that is exhibited by the linear sweep methods. At the same time, the number of turns does not increase in the presence of obstacles, thus adding a desirable feature to the coverage path attained.

## Assumptions:

It is essential that the following assumptions be taken into consideration before the experiments and the respective results can be described:

- The obstacles are circular with radius and centers randomized. - the obstacles do not touch the boundary at more than one point.

- The upper and lower constraints on the sweep lines are given by the previously traced path and the segment along which the next sweep is expected - this allows minimum overlap of successive trajectories.

- Since the coverage radius is constant and small compared to the dimensions of the area, the total area covered by the sweep-line trajectory is a function of the path covered by the trajectory - however, since the presence of obstacles causes the line to curve around the obstacle, an effective measure of the performance of the algorithm is to consider the ratio of total path covered with no obstacles and that covered with obstacles - this allows us to determine the trajectory covers the area in as small as a path as possible.

- Now, for the generated trajectory - only the sweep line cost and not the turning cost is considered for minimization - because in most coverage applications, the system dynamics change drastically when the robot is turning - moreover, the calculation of turning cost generally depends on the inertial parameters of the robots - since only a point robot is considered in this case, the calculation of turning cost in our case becomes futile. To counter this and also considering that empirically the turning cost is much higher than the sweep-line cost, the approach followed by all the generic coverage algorithms is adopted - using different criteria (MSA in this case) to ensure minimum number of turns during the sweeping process.

# EXPERIMENTS AND ANALYSIS

The described algorithm was tested in a MATLAB environment using the TOMLAB-PROPT software. Note that this software makes use of direct collocation methods for obtaining the solution for optimal control problems - although the solution has been analytically obtained by using the Pontryagin's Min-Max principle, the performance evaluation of the algorithm required considering a large number of obstacle constraints, hence a software using pseudospectral optimal control approach was utilized to construct the solutions - the similarity in the solutions obtained from both the methods has been verified. The code is available at https://github.com/Ankitvm/Coverage_Path_Planning-.git

The algorithm was tested for its effectiveness in two cases : first considering an area with no obstacles and then considering an area with a fixed number of obstacles. Moreover, the variation in time, energy and total path covered with respect to parameter variations, namely, variation of size of obstacles, number of obstacles and the weight were also considered.

### Case 1 : Fixed Area With No Obstacles:

The simulation for the performance with no obstacles was carried out by considering a convex polygonal area with fixed dimensions for area coverage. In this case, a rectangular field with dimensions 10 x 10 sq. units was utilized. Since the field was a right quadrilateral, the optimal directions of sweep were found to be 0° and 90°. A horizontal sweep direction was selected for generated the sweeping pattern. The coverage path obtained as a result of the simulation has been illustrated in Fig.(9). Also the curves for the area covered, energy and time taken as functions of the iterations have been plotted in Fig.(10).

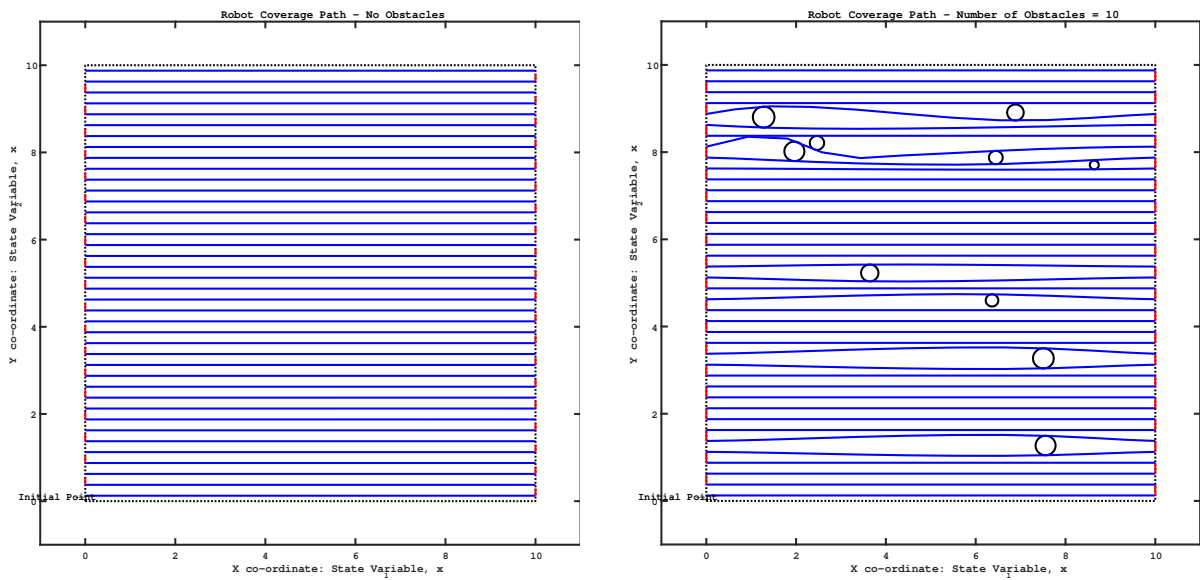

Figure 9: Fixed area with no obstacles - Plot for Robot coverage

Figure 10: Fixed area with 10 obstacles - Plot for Robot coverage

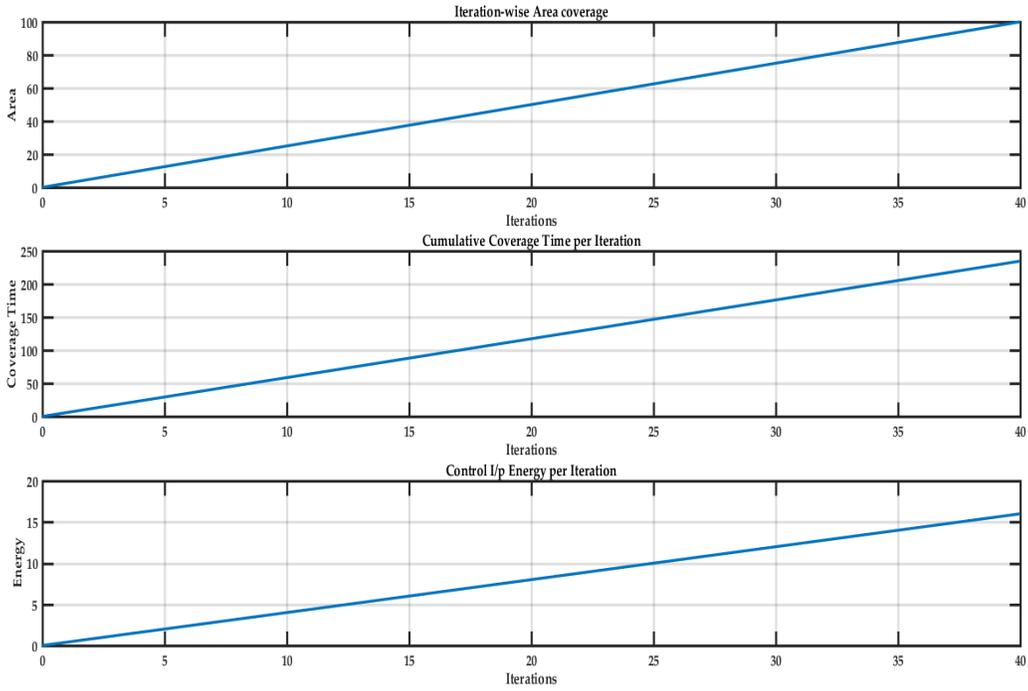

Figure 11: Fixed area with no obstacles - Plot for coverage area, time and energy

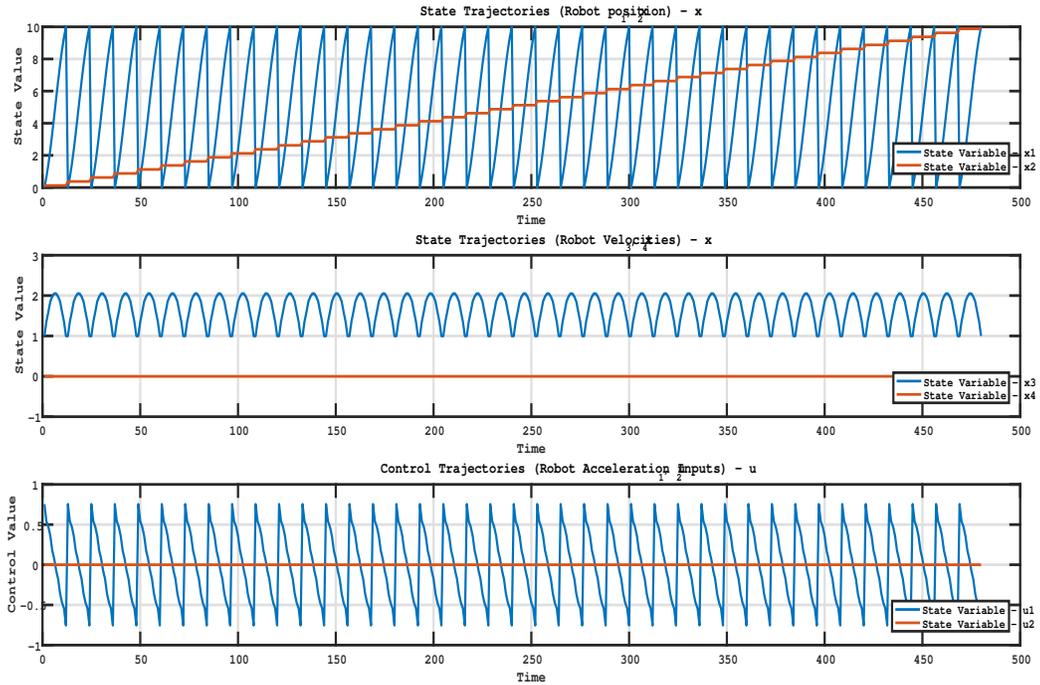

Figure 12: Fixed area with no obstacles - Plot for state and control trajectories

**Observations and Comments:**

From Fig.(9), it is evident that in the absence of obstacles, the algorithm generates a line sweep pattern much similar to a boustrophedon or trapezoidal pattern acting on a convex polygon - this ensures that the sweep pattern retains the desirable characteristics of the mentioned decomposition algorithms in the absence of obstacles. This retention is further confirmed from the plots of time, energy and area covered in Figs. (11.a.),(11.b.),(11.c.) where all the three variables maintain almost a linear relationship with the iterations.

### Case 2 : Fixed Area with Random Obstacles:

Following the simulation with no obstacles, the algorithm simulation was carried out for the same fixed area with the same dimensions but with the presence of obstacles. The simulation environment generated a fixed number of disconnected obstacles with random locations and random radii that are disconnected, i.e, not touching the boundary at more than one point[4]. An illustration of the coverage path generated and the performance plots is shown in Figs (11.a.),(13),(14) respectively.

### Observations and Comments:

From Fig.(12), it can be seen that each sweep is an optimal trajectory serving as a solution to the minimum time/energy optimal control problem in the presence of obstacles. Now for sections of the area where the obstacles are absent the sweep assumes the parallel straight-line nature of a line-sweep based based pattern while in the vicinity of the obstacles, the trajectory curves around the obstacles while maintaining minimum time/energy optimality. This allows the sweep patterns to carry out coverage planning for obstacles of considerable size without having to resort to cellular decomposition. While optimal trajectory is no longer strictly straight parallel, the nature trajectory still maintains little overlap with the adjacent sweeps. This is evident from the curves for the area, time and energy plots in Figs. (13.a.),(13.b.),(13.c.) wherein the plots are almost to the previous case with no obstacles - only changing trajectories when an obstacle lies in the sweep path.

### Performance Evaluation - Response to Parameter Variation:

The performance of the algorithm with respect to the changes in different parameters was carried to determine the robustness of the algorithm under these variations. The three parameters under consideration were the weight (w) for defining the Lagrangian cost, the size of obstacles in the field and the number of obstacles in the environment. The response of each has been illustrated and commented upon ahead:

### Change in Weight, w:

The plot showing the trends in total energy $E$, total coverage time $t_{f_{tot}}$ and area covered $A_{tot}$ has been shown in Fig (15) - as can be seen from the plot, as the weight varies from 0.1 to 0.9, the total control input energy required $||u_1(t)^2 + u_2(t)^2||^2$ shows an increasing trend whereas the total coverage time reduces with increasing $w$. Since the purpose of the weight factor is to determine the priority for minimum time or minimum energy operation for each sweep, we see that its effect on the total time and energy is additive. On the other hand, the change in total area covered is negligible which indicates that $w$ does not affect the area coverage to a veritable degree.

### Number of Obstacles, $N_{obs}$:

The plot showing the trends in total energy $E$, total coverage time $t_{f_{tot}}$ and path coverage ratio $A_{rel}$ has been shown in Fig.(16) - the plots indicate that the algorithm shows a consistent performance for a good range of the number of obstacles. It should be noted that for the determining the response to variation in obstacle parameters, the ratio of the total coverage with obstacles and without obstacles is considered. This is because although the trajectory covers almost an equivalent amount of area for any case, the efficiency of the trajectory generated can be more concretely determined by sweeping the same area through as short a path as possible. Since the shortest possible path is nothing but a straight line which is resultant trajectory for a fixed area with no obstacle, the efficiency can thus be effective determined by considering the ratio of the path covered with obstacles to that covered without obstacles.

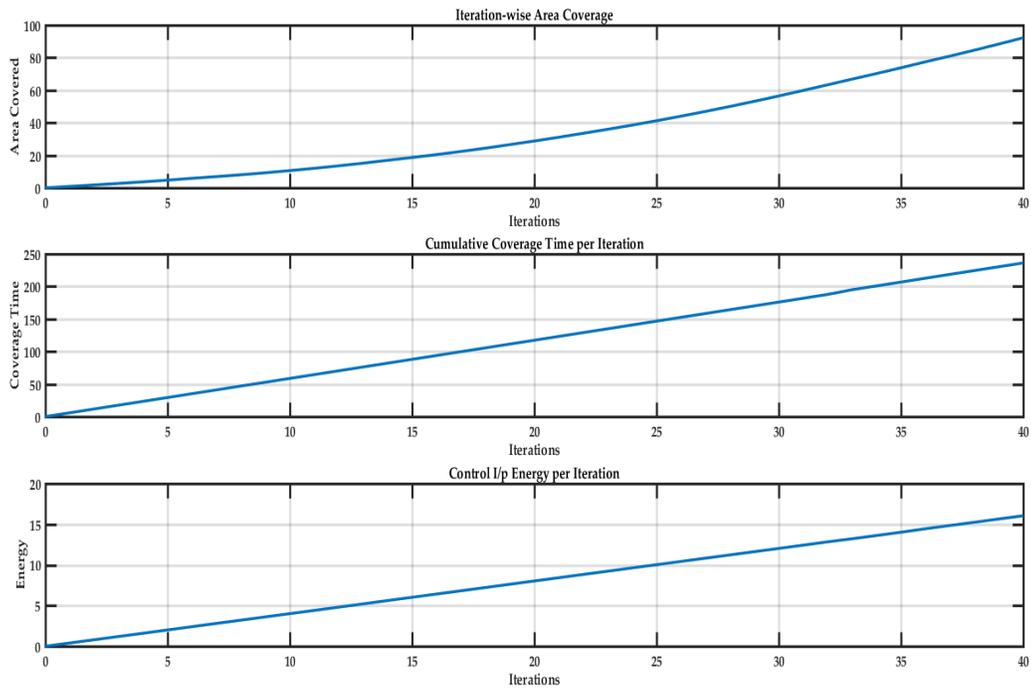

Figure 13: Fixed area with 10 obstacles - Plot for coverage area, time and energy

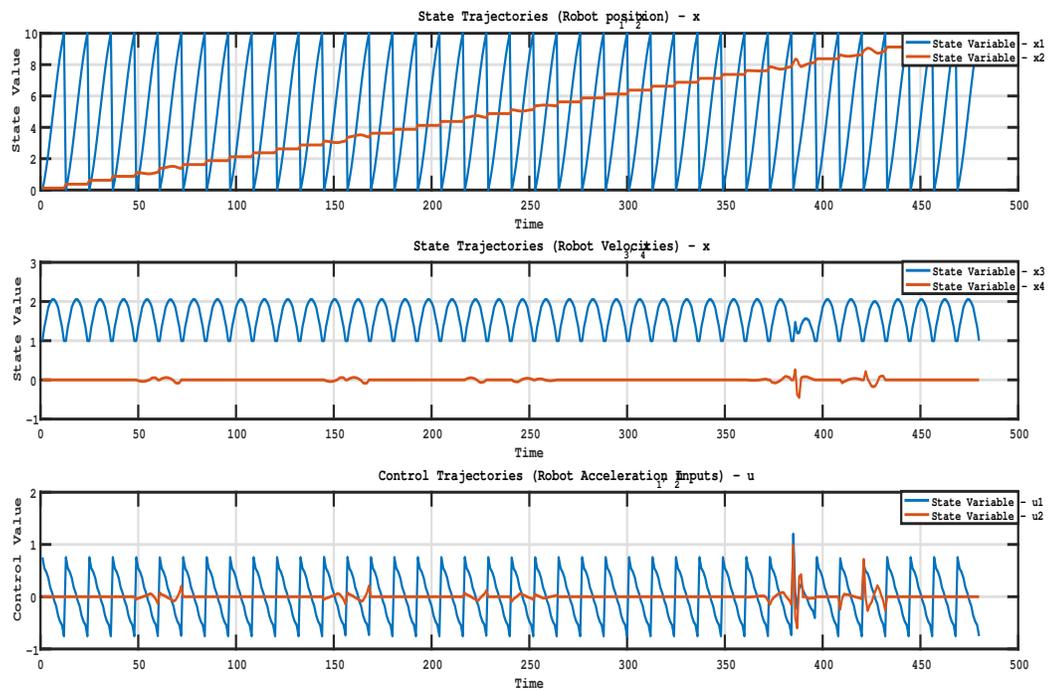

Figure 14: Fixed area with 10 obstacles - Plot for state and control trajectories

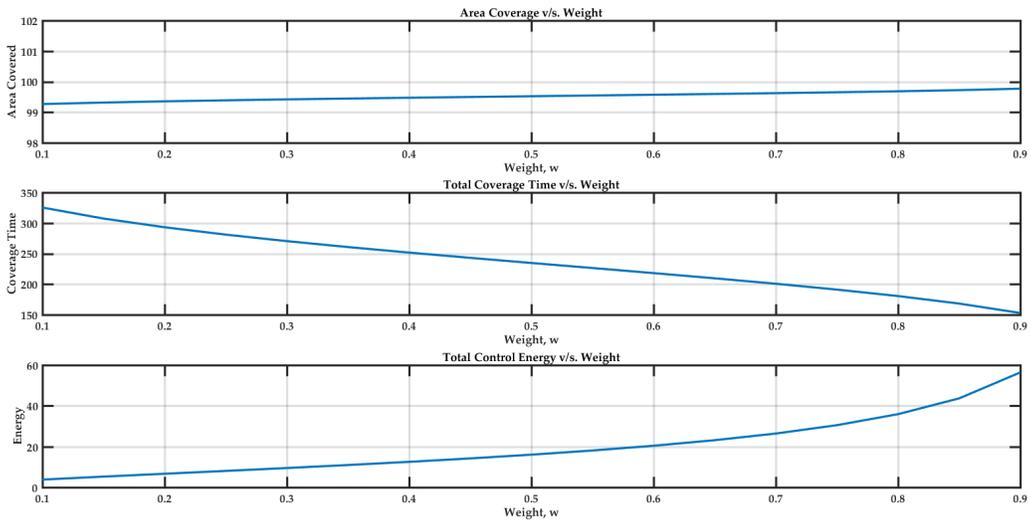

Figure 15: Response to parameter variation - weightage(w)

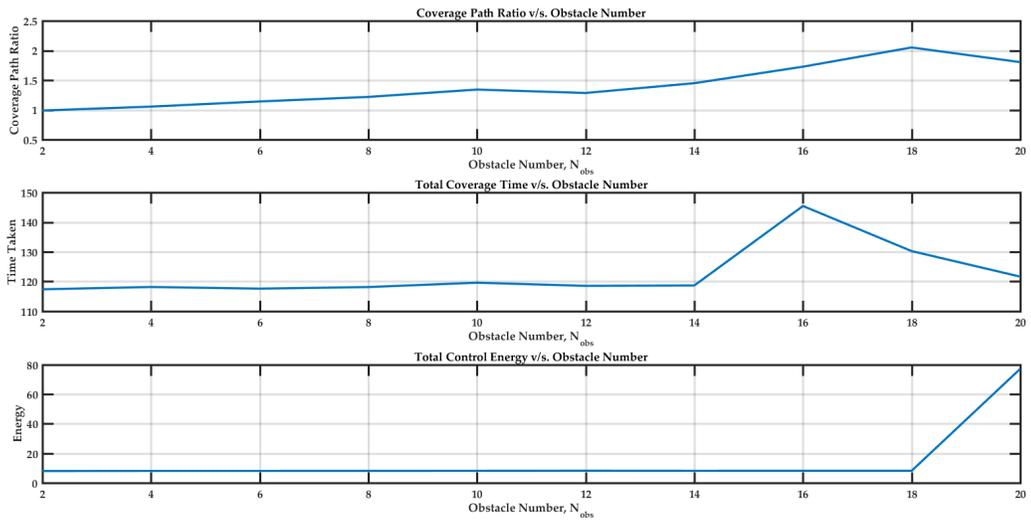

Figure 16: Response to parameter variation - obstacle number

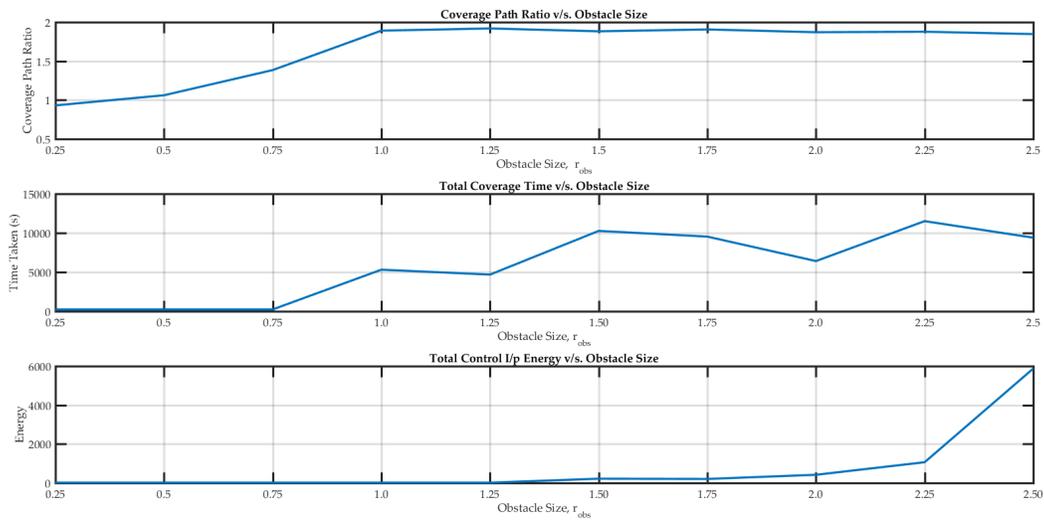

Figure 17: Response to parameter variation - obstacle size

**Size of Obstacles, $r_{obs}$:**

The plot showing the trends in total energy $E$, Total coverage time $t_{f_{tot}}$ and path coverage ratio $A_{rel}$ has been shown in Fig.(17) - The above graph strongly indicates the sensitivity of the algorithm to the size of the obstacle. This should be evident from the fact the sweep lines are primarily constrained to a reduced section to maintain minimum overlap with neighboring trajectories - hence, an obstacle size greater than the constraint width can very likely present contradicting constraints to the optimal control problem.

# CONCLUSION

From the hypotheses presented, modified costs formulated and the simulations that followed, it can be concluded the algorithm for optimal area coverage presents a satisfactory response with regard to the assumptions and constraints specified in the problem formulation. The results of the implementation indicate that the area coverage is independent of the weight factor $w$, utilized in the minimum time/energy cost and is also relatively optimal for a considerable number of obstacles. Observing that the performance degrades for increasing obstacle size, this represents a constraint in the proper performance of the algorithm - thus, to modify the algorithm to cater to larger obstacles with minimum overlap is the subject of future work for this project. Other possible scopes for future work include coverage in a dynamic environment and developing a closed form expression for area coverage cost.

### Workload

**Debasmit:** Pseudo-Spectral Control, Cost Formulation/Analysis    **Pranay:** Heuristic Planning,Cost Modification and Simulation    **Ankit:** Algorithm - Development,Simulation and Implementation